\documentclass[10pt,conference,letterpaper]{IEEEtran}

\IEEEoverridecommandlockouts

\pdfoutput=1		

\usepackage{cite}
\usepackage{amsmath,amssymb,amsfonts}
\usepackage{booktabs}
\usepackage{graphicx}

\usepackage{textcomp}
\usepackage{xcolor}
\usepackage{bm}
\usepackage{float}
\usepackage{algorithm}
\usepackage {algpseudocode}
\usepackage{algorithmicx}
\usepackage{wrapfig}
\usepackage{balance}
\usepackage{caption}
\usepackage{subcaption}
\usepackage{colortbl}
\usepackage[normalem]{ulem}
\usepackage{comment}
\usepackage{multirow}

\def\endthebibliography{%
	\def\@noitemerr{\@latex@warning{Empty `thebibliography' environment}}%
	\endlist
}
\def\BibTeX{{\rm B\kern-.05em{\sc i\kern-.025em b}\kern-.08em
		T\kern-.1667em\lower.7ex\hbox{E}\kern-.125emX}}

\usepackage{titling}
\renewcommand\maketitlehooka{
  \setlength\parindent{0pt}
  \begin{minipage}{\textwidth}
    \center
    This work has been submitted to the IEEE for possible publication.\\
    Copyright may be transferred without notice, after which this version may no longer be accessible.
  \end{minipage}
  \vspace{0.2in}
}
\date{}	

\begin{document}

\title{Geographical Node Clustering and Grouping\\to Guarantee Data IIDness in Federated Learning}

\author{Minkwon Lee, Hyoil Kim, {\it Senior Member, IEEE}, Changhee Joo, {\it Senior Member, IEEE}
\thanks{Minkwon Lee and Hyoil Kim are with the Department of Electrical Engineering, the Ulsan National Institute of Science and Technology (UNIST), Ulsan 44919, Republic of Korea (e-mail: \{minkwon1114,hkim\}@unist.ac.kr).}
\thanks{Changhee Joo is with the Department of Computer Science and Engineering, Korea University, Seoul 02841, Republic of Korea (email: changhee@korea.ac.kr).}
\thanks{Hyoil Kim is the corresponding author.}
}

\pagenumbering{gobble}	

\maketitle

\begin{abstract}
Federated learning (FL) is a decentralized AI mechanism suitable for a large number of devices like in smart IoT.
A major challenge of FL is the non-IID dataset problem, originating from the heterogeneous data collected by FL participants, leading to performance deterioration of the trained global model.
There have been various attempts to rectify non-IID dataset, mostly focusing on manipulating the collected data.
This paper, however, proposes a novel approach to ensure data IIDness by properly clustering and grouping mobile IoT nodes exploiting their geographical characteristics, so that each FL group can achieve IID dataset.
We first provide an experimental evidence for the independence and identicalness features of IoT data according to the inter-device distance, and then
propose \textit{Dynamic Clustering} and \textit{Partial-Steady Grouping} algorithms that partition FL participants 
to achieve near-IIDness in their dataset while considering device mobility. 
%
Our mechanism significantly outperforms benchmark grouping algorithms at least by 110 times in terms of the joint cost between the number of dropout devices and the evenness in per-group device count, with a mild increase in the number of groups only by up to 0.93 groups.
\end{abstract}


\section{Introduction}
\label{sec:intro}

Federated Learning (FL) is a promising AI technology for providing intelligent services to resource-limited devices in diversified mobile environments (e.g., IoT, autonomous vehicles, smartphones).
Centralized AI service incurs privacy concerns in personalized services like smart healthcare \cite{Baker2017Access} and smart home \cite{Malche2017ISMAC}, due to the transmission of local device data to the central server.
On the contrary, FL is a decentralized technology that can effectively address such issues, by conducting local learning on end-devices using their own data and transmitting only the trained models to the server \cite{Lim2020CST}.

Nevertheless, FL suffers from the problem of non-IID training data which is originated from the heterogeneity of collected mobile data among the devices experiencing different task-operating environments. 
Specifically, the non-IID problem implies that when the distribution of each device's collected data is either correlated or non-identical to each other, it may cause a degraded performance (in terms of test accuracy) of the federated model compared to the one trained with IID datasets \cite{Zhao2018Arxiv}.
In addition, it may incur slower convergence of FL, leading to consuming more communication and computational overhead in the overall operation of FL.

In the literature, there have been several work proposing various mechanisms to alleviate the negative impact of non-IID data on the FL performance.
Among them, some studies focused on resolving data heterogeneity through algorithmic approaches as follows.
\cite{Smith2017NeurlPS} proposed integration of FL with multi-task learning which tries to learn the relationship among similarly-structured local models to improve the performance of global model.
In \cite{Wang2020INFOCOM}, participant selection via deep Q-learning was proposed
to select the best devices that can maximize the validation accuracy per training round.
Alternatively, there also exist data sharing strategies to tackle the problem as follows.
\cite{Zhao2018Arxiv} proposed sharing 
the global data from the federated server with its clients, in order to reduce the model's weight divergence by performing local training with local and global dataset combined together.
Similarly, \cite{Itahara2021TMC} proposed a strategy of training the local model in two stages, once with local data and again with globally-shared data.
%
None of the above, however, exploited the nature of communication environments in developing the FL techniques to restore data IIDness.
Furthermore, it is hard to find work applicable to time-varying data distribution due to node mobility.

This paper focuses on the geographical aspect of mobile data to develop a novel FL mechanism to achieve near-IIDness of the IoT data.
%
Smart IoT devices equipped with multi-modal sensors (e.g., 
camera, microphone, thermometer) may run federated AI algorithms to perform smart tasks, 
such as autonomous robots in a warehouse monitoring the environment (e.g., heat signature) or
managing inventory (e.g., with vision sensors) \cite{Konstantinidis2022MED, Ho2024TASE}, and 
a swarm of autonomous drones with vision sensors collaborating for surveillance tasks \cite{Ding2018MCOM}.
In such scenarios, nearby devices could obtain correlated sensor readings (e.g., temperature, images taken toward similar directions), while such correlation diminishes 
as they get farther away.
When they are too far apart, however, their collected data may follow non-identical distributions (e.g., temperature near the entrance vs. at the center of the building).
%
%

In such a vein, we propose mobile node clustering and grouping techniques to achieve almost-IID data distribution in FL.
%
Motivated by 
aforementioned insights, 
we developed node clustering and grouping algorithms that takes into account node mobility and inter-node correlation and distributional similarity of their IoT data, utilizing graph coloring to satisfy the data IIDness among the nodes in the same FL group.
To the best of our knowledge, this is the first work exploiting geographical features of IoT data to demonstrate its potential for resolving the issue of non-IID dataset in the FL training.

Our contributions are four-fold, as summarized as follows.
\begin{itemize}
\item Our experimental study in Section~\ref{sec:experiment} demonstrates the geographical effect on the IIDness of collected data, 
revealing that the distance between data collection locations plays a crucial role in the geographical characteristics, along with the surrounding environment. 
\item We propose \textit{Dynamic Clustering}, a geographical node clustering algorithm to achieve near-identical distribution among mobile data by upper-bounding the inter-device distance in a FL cluster.
\item We also propose \textit{Partial-Steady Grouping}, a geographical node grouping algorithm to guarantee near-independence between mobile data by lower-bounding the minimum inter-device distance in each node group.
\item Via extensive simulations, our proposal is compared with state-of-the-art node grouping algorithms. Our algorithm significantly outperforms others in terms of the joint cost between the number of dropped devices from training and the evenness in per-group device count. 
\end{itemize}

The rest of the paper is organized as follows.
Section~\ref{sec:experiment} presents our experimental results as an evidence of the geographical characteristics in IoT data distributions.
Then, Section~\ref{sec:model} introduces the system model of our FL mechanism.
Section~\ref{sec:main algorithm} proposes dynamic clustering and grouping algorithms to restore the IIDness within each FL group, and 
Section~\ref{sec:eval} evaluates the proposed algorithms through extensive simulations.
Finally, the paper concludes with Section~\ref{sec:conclusion}.


\vspace{0.1in}

\section{Experimental Evidence for Geographical Characteristics of IoT Data}
\label{sec:experiment}
This section demonstrates our experimental study on the geographical characteristics of IoT data, to verify our conjecture of the data correlation and identicalness depending on the inter-device distance.
Such a study is necessary since there exists very few work investigating the IIDness feature of IoT data, 
and it is hard to find public datasets with the usual IoT scale (i.e., room to building).
Although \cite{Hsieh2020PMLR} studied the variation in statistical characteristics of the geographically-collected dataset, the collection locations were sampled at the scale of continents thus not suitable for revealing the geographical characteristics of IoT data distribution. 

\begin{figure}[!t]
    \centering
    \includegraphics[height=3.8cm]{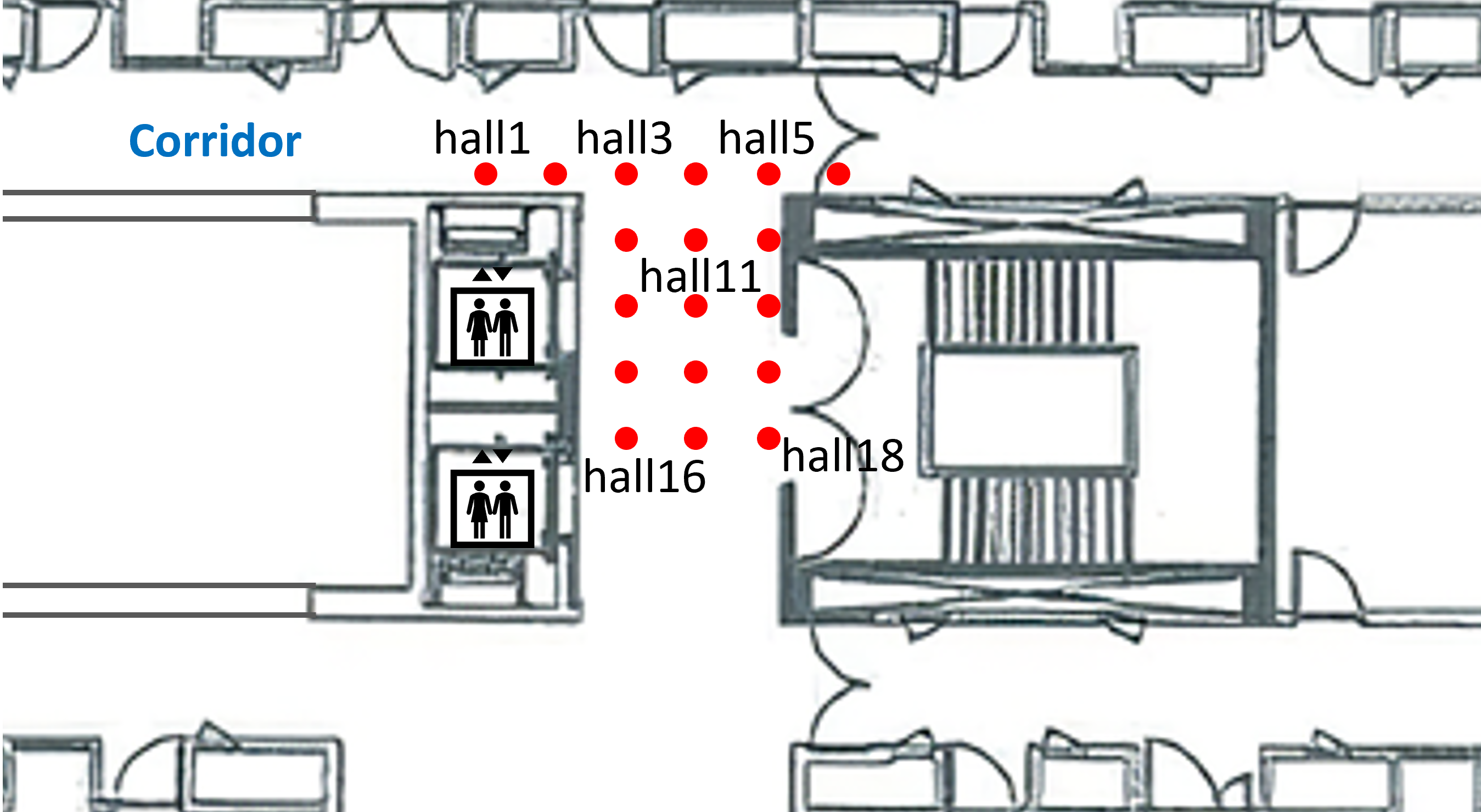}
    \caption{Experimental environment and IoT locations}
    \label{fig:experiment_env}
\end{figure}

In the experiment, we measured temperature at various indoor locations, as an exemplifying case of IoT data in smart healthcare, smart buildings, etc.\footnote{Since this is the first study on the geographical IoT data features, we plan to diversify the IoT data types in the future experiments.}
Fig.~\ref{fig:experiment_env} illustrates 18 measurement locations (labeled as {\it hall1} through {\it hall18}) in our campus building where two adjacent locations are 1.35 meters apart.
The considered area consists of an elevator hall, a staircase on its right, and a long corridor above the hall.\footnote{We have not tested the locations beyond {\it hall1} (to the left) and {\it hall5} (to the right) due to the consistent temperature deeper in the corridor. In addition, we excluded the lower half area of the tested environment due to symmetry.}
Then, we concurrently measured the temperature at these points using multiple thermometers, which had been repeated 20 times at various times and days. 
Since the measurements were performed in late November, the data points closer to the staircase tend to have lower temperature. 

\subsection{Geographical Characteristics on the Identicalness}

\begin{figure}[!t]
    \centering
    \includegraphics[width=\linewidth]{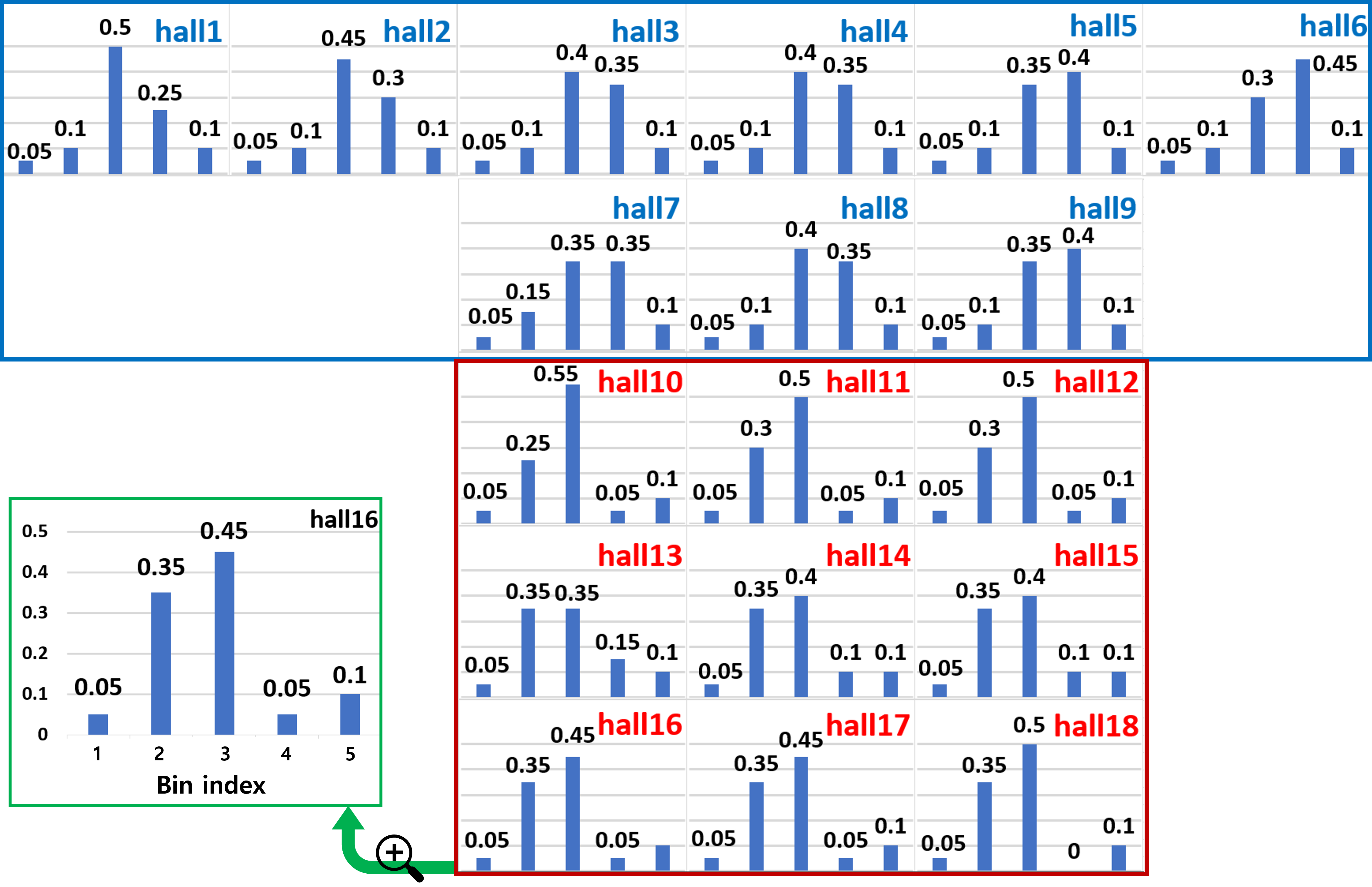}
    \caption{Temperature data distribution for each location}
    \label{fig:histogram}
\end{figure}

With collected data, we compared per-location histograms to investigate the data identicalness feature.
To generate each per-location histogram, we considered five bins where the bin width is given as the `maximum difference in temperature over the entire dataset' divided by five. 
Each bin's frequency of occurrence is obtained as the number of data points belonging to each bin divided by the total number of measurements (i.e., 20), producing a probability distribution. 
Fig.~\ref{fig:histogram} depicts thus-constructed histograms, which have been arranged according to the measurement locations in Fig.~\ref{fig:experiment_env}.
After obtaining the histograms, we compared the distribution of the reference data point `hall18' and that of another data point by using the KL-divergence \cite{Kullback1951Math.Stat.}, which
measures the difference between the two probability distributions.
Table~\ref{tab:kl_div} depicts thus-constructed KL values arranged by the corresponding locations, where a larger value implies that the distributions are less identical.

\begin{table}[!t]
\centering
\footnotesize
\begin{tabular}{cccccc}
\hline
\multicolumn{6}{c}{\textbf{KL value}} \\ \hline
\rowcolor[HTML]{96FFFB}
  0 (hall1) &
  0.0073 &
  {\color[HTML]{000000} 0.0285} &
  {\color[HTML]{000000} 0.0285} &
  {\color[HTML]{000000} 0.0632} &
  \cellcolor[HTML]{45EFE8}{\color[HTML]{000000} 0.1113} \\
\rowcolor[HTML]{F0F0F0}
{\color[HTML]{000000} } &
  {\color[HTML]{000000} } &
  \cellcolor[HTML]{96FFFB}{\color[HTML]{000000} 0.0537} &
  \cellcolor[HTML]{96FFFB}{\color[HTML]{000000} 0.0285} &
  \cellcolor[HTML]{96FFFB}{\color[HTML]{000000} 0.0632} &
  {\color[HTML]{000000} } \\
{\color[HTML]{000000} } &
  {\color[HTML]{000000} } &
  \cellcolor[HTML]{45EFE8}{\color[HTML]{000000} 0.1486} &
  \cellcolor[HTML]{45EFE8}{\color[HTML]{000000} 0.2491} &
  \cellcolor[HTML]{45EFE8}{\color[HTML]{000000} 0.2491} &
  {\color[HTML]{000000} } \\
\rowcolor[HTML]{F0F0F0}
{\color[HTML]{000000} } &
  {\color[HTML]{000000} } &
  \cellcolor[HTML]{45EFE8}{\color[HTML]{000000} 0.2370} &
  \cellcolor[HTML]{46E8E1}{\color[HTML]{000000} 0.2576} &
  \cellcolor[HTML]{46E8E1}{\color[HTML]{000000} 0.2576} &
  {\color[HTML]{000000} } \\
{\color[HTML]{000000} } &
  {\color[HTML]{000000} } &
  \cellcolor[HTML]{46E8E1}0.3106 &
  \cellcolor[HTML]{46E8E1}0.3106 &
  \cellcolor[HTML]{36DAD4}0.4319 &
  {\color[HTML]{000000} } \\ \hline
\end{tabular}
\\
\begin{tabular}{cccccc}
\hline
\hline
\cellcolor[HTML]{14C7CB}1.2561 &
  \cellcolor[HTML]{1CC0C4}1.5393 &
  \cellcolor[HTML]{24BABE}{\color[HTML]{000000} 1.8365} &
  \cellcolor[HTML]{24BABE}{\color[HTML]{000000} 1.8365} &
  \cellcolor[HTML]{329A9D}{\color[HTML]{000000} 2.1472} &
  \cellcolor[HTML]{34696D}{\color[HTML]{000000} 2.4712} \\
\rowcolor[HTML]{F0F0F0}
{\color[HTML]{000000} } &
  {\color[HTML]{000000} } &
  \cellcolor[HTML]{24BABE}{\color[HTML]{000000} 1.7990} &
  \cellcolor[HTML]{24BABE}{\color[HTML]{000000} 1.8365} &
  \cellcolor[HTML]{329A9D}{\color[HTML]{000000} 2.1472} &
  {\color[HTML]{000000} } \\
{\color[HTML]{000000} } &
  {\color[HTML]{000000} } &
  \cellcolor[HTML]{35E8E1}{\color[HTML]{000000} 0.1650} &
  \cellcolor[HTML]{35E8E1}{\color[HTML]{000000} 0.1504} &
  \cellcolor[HTML]{35E8E1}{\color[HTML]{000000} 0.1504} &
  {\color[HTML]{000000} } \\
\rowcolor[HTML]{F0F0F0}
{\color[HTML]{000000} } &
  {\color[HTML]{000000} } &
  \cellcolor[HTML]{37D5CF}{\color[HTML]{000000} 0.6275} &
  \cellcolor[HTML]{36DAD4}{\color[HTML]{000000} 0.3721} &
  \cellcolor[HTML]{36DAD4}{\color[HTML]{000000} 0.3721} &
  {\color[HTML]{000000} } \\
{\color[HTML]{000000} } &
  {\color[HTML]{000000} } &
  \cellcolor[HTML]{96FFFB}0.1491 &
  \cellcolor[HTML]{96FFFB}0.1491 &
  \cellcolor[HTML]{96FFFB}0 (hall18) &
  {\color[HTML]{000000} } \\ \hline
\end{tabular}%
\caption{Per-location KL-divergence} 
\label{tab:kl_div}
\end{table}

In Fig.~\ref{fig:histogram}, the histograms in the red box are skewed to the left due to the influence of cold airflow from the staircase, whereas the ones in the blue box present higher temperatures on average.
Moreover, the distribution shapes in the same colored box look more similar to each other while the inter-color shapes present notable difference (i.e., the ones in the red box tend to be skewed to the left whereas the ones in the blue box are skewed to the right).
%
Similarly, the KL value in Table~\ref{tab:kl_div} tends to increase fast as the location gets farther from the reference point in the vertical direction, presenting a significant discrepancy when crossing the border between blue and red areas. 
In addition,
the KL value of adjacent locations tends to be similar especially when the two locations are placed horizontally, which seems to be caused by the structural difference between the corridor and the elevator hall.

As a result, it appears that {\it there exists the maximum distance between two devices below which their data distributions satisfy near-identicalness},
which will be denoted by $d_{max}$ in the sequel.
%
In practice, the mobile network or FL service provider can pre-determine $d_{max}$ by deploying probing devices in the service area and utilizing their collected data.
Specifically, $d_{max}$ can be derived by first calculating KL divergence values between a reference device and others, and then determining the inter-device distance below which the KL value stays smaller than a pre-defined threshold with a target confidence (e.g., 95\%), where the distance becomes $d_{max}$.

\subsection{Geographical Characteristics on the Independence}

We examined the correlation between measurement points by computing the Pearson correlation coefficient $\rho_{X,Y}$ between the reference point $X$ and a point $Y$ \cite{Pearson1895Notes}, while varying $X$ as one of three candidates.
Denoting their measurements respectively by $\{ X_i \}$ and $\{ Y_i \}$, $i = 1, ..., 20$, $\rho_{X,Y}$ is given as $\rho_{X,Y} = {cov(X,Y)}/{\sigma_X \sigma_Y}$
where ${\sigma_X}^2$, ${\sigma_Y}^2$ are the unbiased sample variance and $cov(X,Y)$ is the sample covariance.

\begin{figure}[!t]
  \centering
  \includegraphics[width=0.8\columnwidth]{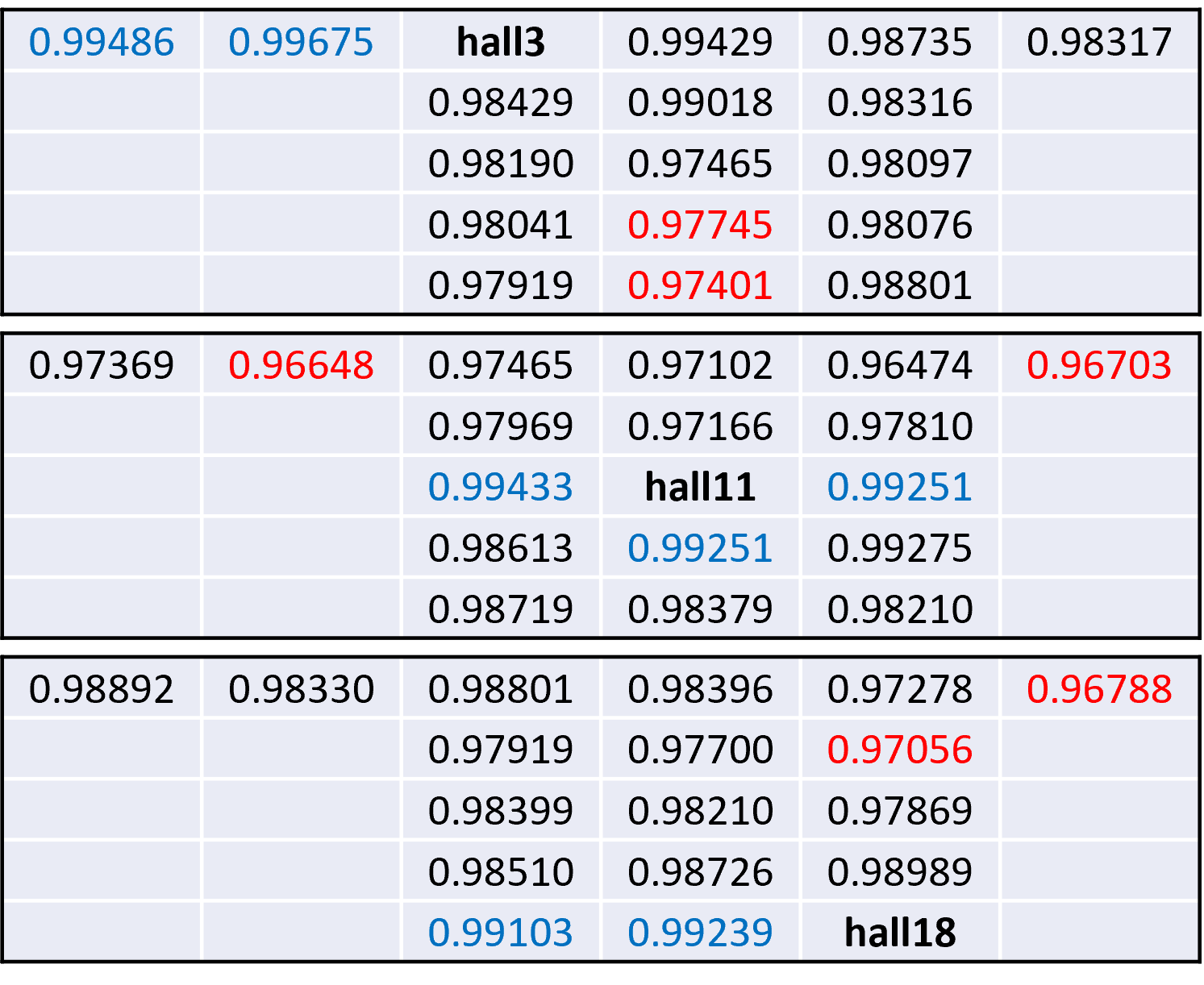}
  \captionsetup{justification=centering}
  \caption{Correlation coefficient with varying reference points (in bold). The largest (or smallest) values are in blue (or red)}
  \label{fig:correlation_coefficient_table}
\end{figure}

Fig.~\ref{fig:correlation_coefficient_table} illustrates
the results with three reference points: {\it hall3}, {\it hall11}, {\it hall18}.
Noticeably, the largest values in blue are close to the reference point, while the smallest ones are distant from it.
In the meantime, all the values being quite large suggests high consistency in temperature fluctuations over time in the  environment, which is natural for indoor thermal data.
Nevertheless, upon increasing the distance from the reference point, a noticeable trend is observed wherein $\rho_{X,Y}$ exhibited a decline indicating that closer data points are more correlated.

Therefore, the experiment revealed that the IoT data also exhibits geographical characteristics regarding data independence.
The observed tendency of correlation values with varying inter-device distance implies that {\it there exists the minimum distance beyond which the data distribution of any two devices may satisfy near-independence},
which will be denoted by $d_{min}$ in the sequel.
%
Similar to $d_{max}$, $d_{min}$ can be pre-determined by calculating $\rho_{X,Y}$ between the measurements of two probing devices, while gradually decreasing the inter-device distance.
Once the correlation becomes larger than a pre-defined threshold with a target confidence (e.g., 95\%) for a certain tested inter-device distance, $d_{min}$ is set as the distance.

\section{System Model}
\label{sec:model}
This work leverages the most popular FL model `FedAvg' \cite{Mcmahan2017PMLR} 
and evolves it to propose our geographical FL mechanism. 


In FedAvg, there exists a single FL server to control $N$ clients/devices which are participating in the learning process.
Initially, the FL server owns the global model $w_G$ and randomly selects a certain number of its clients.
Then, the FL system follows the following steps:
\begin{enumerate}
    \item The server sends the current global model to the selected clients, and client device $i$ locally trains the received model with local dataset $D_i$ such as $w_i \leftarrow w_i - \eta \Delta L(w_i;D_i)$ where $w_i$ is the local model's parameter, $\eta$ is the learning rate, and $\Delta L$ is the gradient of the loss function \cite{Lim2020CST}. $D_i$ may be split into several batches.
    \item When device $i$ finishes updating its parameter $w_i$, it transmits the updated model back to the FL server. Then the FL server aggregates the received $N$ local models via averaging, such as $w_G\leftarrow \frac{1}{N} \sum_{i=1}^N{w_i}$.
\end{enumerate}
If the current global model fails to converge,
the FL server repeats the above steps by re-selecting the clients.
In this work, a single iteration consisting of steps 1 and 2 is called a \textbf{round}.

\begin{figure}[!t]
    \centering
    \includegraphics[width=0.9\columnwidth]{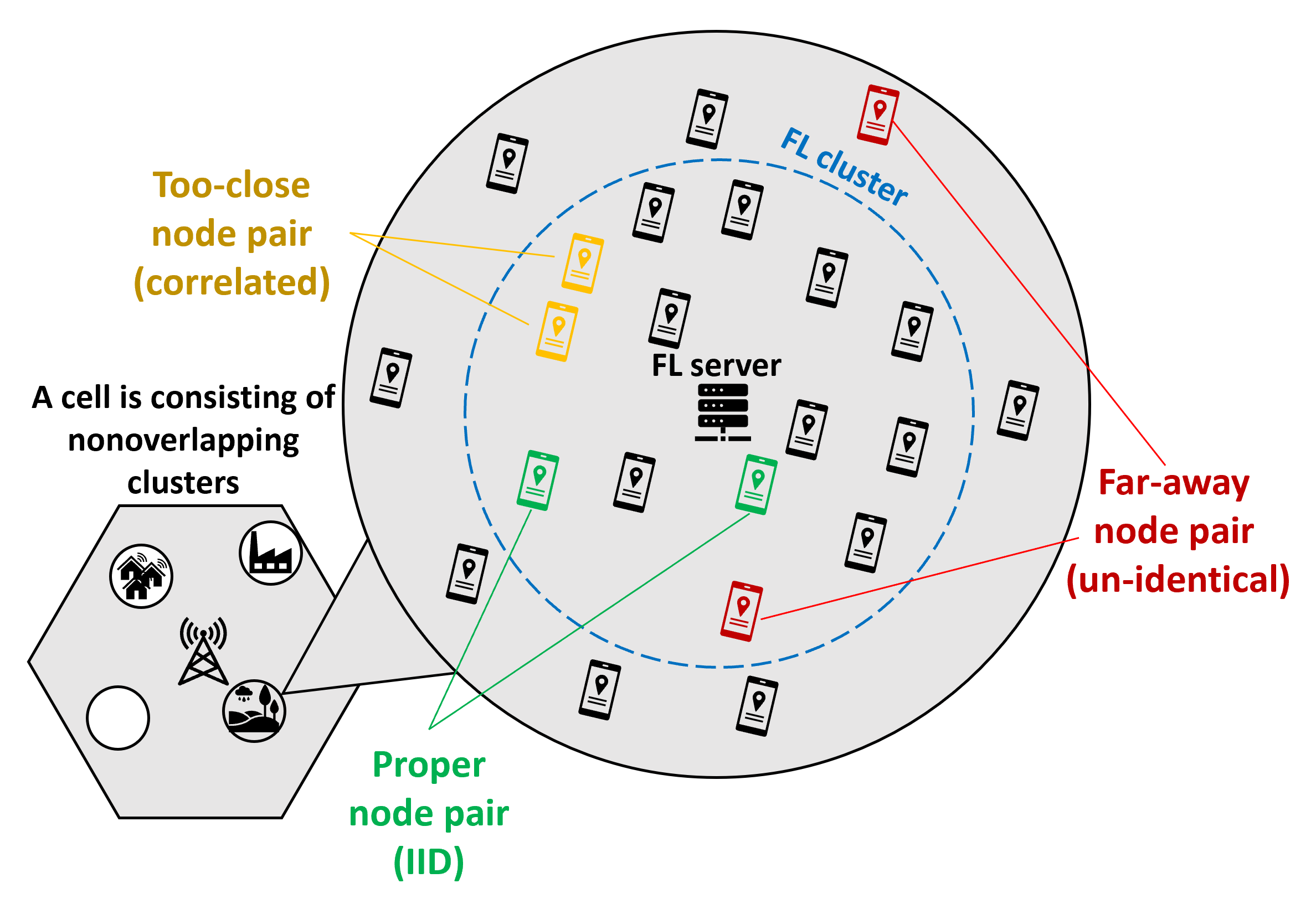}
    \caption{The concept of an FL cluster}
    \label{fig:cluster}
	\vspace{-0.1in}
\end{figure}

Then, to construct our FL model, we introduce supplemental concepts evolved from FedAvg.
%
First, we define a single FL system as an \textbf{FL cluster}, comprising one FL server and multiple IoT devices.
Assuming a cell managed by its Base Station (BS) as in Fig.~\ref{fig:cluster}, there could exist multiple intra-cell FL servers each managing a separate FL cluster.
FL servers can be colocated at the BS (e.g., each running as a container) or placed in the cellular core (e.g., running at a mini data center), and hence IoT devices communicate with FL servers via the BS over wireless links.
We also assume such FL clusters do not overlap with each other, which can be achieved by splitting the cell into disjoint sub-cells (each with a designated FL server).
%

\begin{figure}[!t]
    \centering
    \includegraphics[width=0.99\columnwidth]{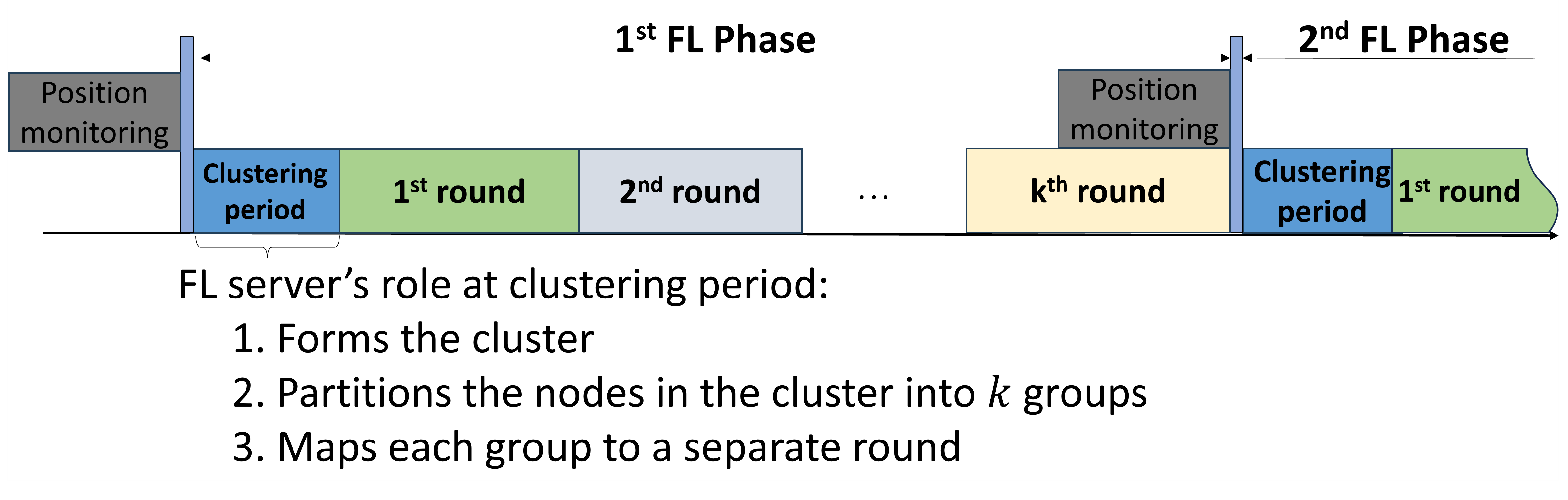}
    \caption{Overview of the proposed FL mechanism}
    \label{fig:FL_mechanism}
    \vspace{-0.05in}
\end{figure}

\begin{figure*}[!t]
    \centering
    \includegraphics[width=0.8\textwidth]{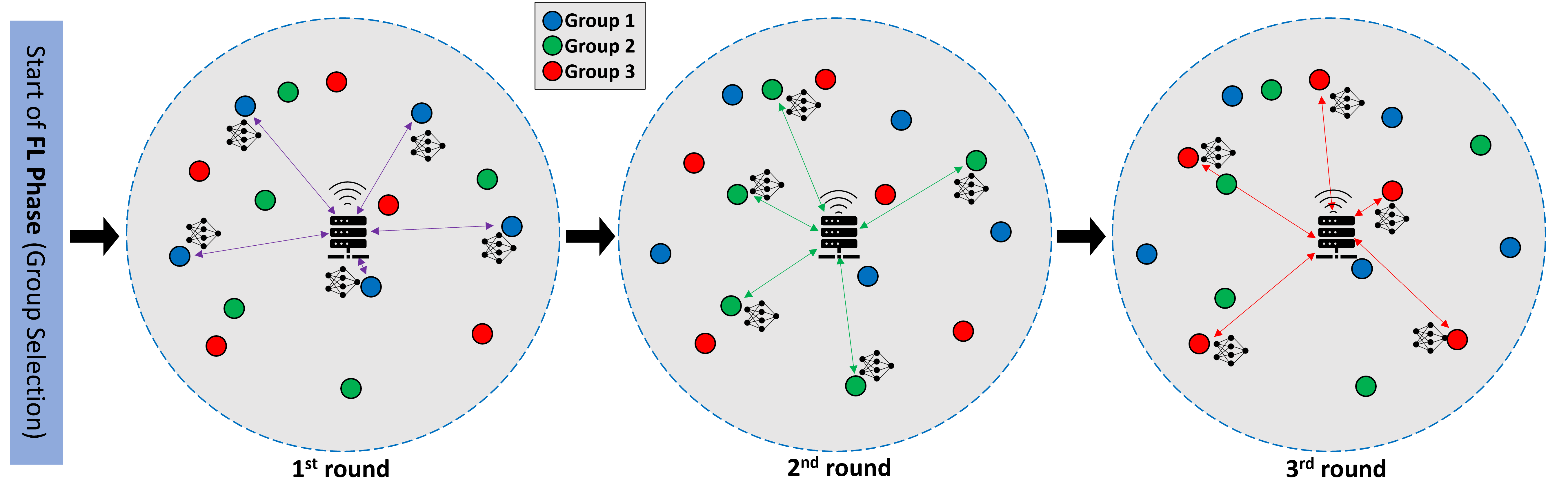}
    \captionsetup{justification=centering}
    \caption{Proposed FL procedure with one group participating per round}
    \label{fig:grps_and_rounds}
	\vspace{-0.05in}
\end{figure*}

The concept of our proposed FL mechanism is illustrated in Fig.~\ref{fig:FL_mechanism}.
There exist a series of \textbf{FL phases}, each of which consists of a clustering period and a sequence of rounds.
Before a new FL phase starts, the BS performs `position monitoring' to collect the position of each nodes in its coverage via location services \cite{3GPP2020R16}, and provides the result to its FL servers.
Based on the location information, the FL server determines the node members of its cluster during the \textbf{clustering period}.
%
Then, the intra-cluster nodes are further partitioned into \textbf{FL groups} (during the same clustering period), where an FL group is formed such that data correlation among the devices in the same group is suppressed by maintaining the minimum separation distance of $d_{min}$ from one another.
For example, Fig.~\ref{fig:grps_and_rounds} illustrates a cluster with three groups, where each group consists of distant nodes so that neighboring nodes may not belong to the same group.
The detailed criterion of grouping is constructed based on the $d_{min}$ value and device mobility, which will be elaborated in Section~\ref{sec:main algorithm}.

In the original \textit{FedAvg}, the clients to participate in FL are randomly selected at the start of each round.
In our model, however, once the intra-cluster nodes are partitioned into FL groups, the FL server assigns each of them to a separate round in advance.
Hence in each round, only the associated group members participate in the FL process (i.e., communicating with the FL server, training local models, etc.) with their own dataset collected during the previous FL phase.
Finally, an FL phase concludes upon the completion of all the rounds.

During an FL phase, the FL server can terminate its learning process if the global model has been converged prior to the end of the phase.
In such a case, the remaining rounds in the phase may be cancelled.
This is reasonable because FL groups are formed to have almost the same statistical properties (thus little to gain further by performing the remaining rounds, once converged), which is achieved by selecting the nodes at least $d_{min}$ apart from each other so that the nodes in each group are almost uniformly populated within the same cluster area.
However, the FL process should resume at the start of the next FL phase, since the network condition could change due to node mobility and environmental variations.

A node that was present during the clustering period could leave the cluster by the time its group period starts due to mobility.
Such a node, however, can still participate in FL unless it leaves the BS's coverage, since the FL server can communicate with the node via the BS.
Although the node should be excluded 
if it has departed from its BS, this happens quite rarely because the cell radius is typically a few kilometers while the time gap between the clustering period and the per-group round is not large enough to incur many missing nodes, considering the limited mobility of IoT devices (e.g., walking speed).
For example, with around ten rounds (typical as will be shown in Section~\ref{sec:eval}), the maximum time gap (of the last round) may not exceed 30 seconds considering a known case where the local online learning at each FL device takes less than 1 second \cite{Luo2021INFOCOM} and hence a round may persist only up to a couple of seconds (including the communication latency with the BS).
Moreover, since later rounds are skipped when the global model converges, the impact of missing nodes would have even less impact.

\section{Dynamic Clustering Algorithm and\\Partial-Steady Grouping Algorithm}
\label{sec:main algorithm}
Assuming a given IoT network has geographical features according to $d_{max}$ and $d_{min}$,\footnote{In case $d_{min} > d_{max}$, our geographical FL method should not be used.}
this section proposes
applying $d_{max}$ to node clustering and $d_{min}$ to node grouping to construct our FL mechanism, 
leveraging on the graph theory.

\subsection{Preliminaries: Graph Theory}

In an undirected graph $G = (V, E)$ with a set of vertices/ nodes $V$ and a set of edges $E$, 
any two distinct vertices connected by an edge are  \textit{adjacent}.\footnote{Note that this work does not consider a loop as an edge.}
The \textit{graph density} of $G$ is the \textit{ratio} of the number of edges in $G$ \textit{to} the maximum possible number of edges (of the complete graph on $V$).
In $G^{\prime}$, 
the \textit{complement graph} of $G$, 
two distinct vertices are adjacent if and only if they are not adjacent in $G$ \cite{Lewis2015Springer}.
In addition, a \textit{clique} is a subgraph of $G$ that has a collection of vertices where every pair of them is connected via an edge.


\subsection{Proposed Clustering Algorithm: Dynamic Clustering}

For an FL system, we form a circular cluster with the diameter of $d_{max}$, within which nodes are treated to have identically-distributed datasets by the definition of $d_{max}$.
The clustering problem, however, is not complete by simply forming a cluster, since node mobility should also be considered.
For instance, a node that has been predominantly located outside the cluster could enter its coverage at the last moment before clustering is performed.
It is also possible that a certain node located at the cluster edge disappears and reappears over time due to high mobility. 
In such cases, the identicalness within the cluster may not be ensured if we blindly let such nodes participate in FL.
As a result, clustering must consider per-node location \textit{history}, instead of just using a snapshot of node locations.

%

In this regard, we propose the `Clustering Suitability' ($CS$) metric, with which only suitable nodes can be identified for guaranteeing intra-cluster identicalness.
For sample time $t \in \{ 1, \ldots, T \}$, $CS$ of device $i$, denoted by $CS(i)$, is defined as:
\begin{eqnarray}
CS(i) = \sum\nolimits_{t = 1}^\textit{T}{ \omega_t \cdot I_{CS}(i,t) } , 
\end{eqnarray}
where the clustering indicator $I_{CS}(i,t)$ is set to 1 when node $i$ is within the cluster at time $t$, and $0$ otherwise. 
In addition, $\omega_t$ is a monotonic non-decreasing discount factor (i.e., $\omega_1 \le \omega_2 \le \cdots \le \omega_T$) to assign larger weight to a more recent location, which satisfies $\sum_{t = 1}^\textit{T} \omega_t = 1$.
Then, over $T$ sampling times, nodes are required to be within the cluster for more than $\xi_{CS}$ (weighted) fraction of time (e.g., 0.7), i.e., $CS(i) \ge \xi_{CS}$, to be eligible for participating in the FL process.
Then, we can obtain the \textit{suitable node set} $\tilde{N}$ ($\subset N$) such as
\begin{eqnarray}
\tilde{N} = \{ i \in N \phantom{0} | \phantom{0} CS(i) \ge \xi_{CS} \} .
\end{eqnarray}

After constructing $\tilde{N}$, the FL server should partition them into a set of groups 
while ensuring that in each group any inter-device distance is larger than $d_{min}$ to achieve near-independence. 
Again, since the mobility of devices may incur variations in the inter-device distance, 
appropriate pairings between such nodes should be established based on their location histories, via another proposed metric called `Pairing Suitability'.
At $t$, if devices $i$ and $j$ are farther than $d_{min}$, the pairing indicator $I_{PS}(i,j,t)$ is set to $1$, and $0$ otherwise.
Then, the Pairing Suitability $PS(i,j)$ between $i$ and $j$ is defined as:
\begin{eqnarray}
PS(i,j) = \sum\nolimits_{t = 1}^\textit{T} \omega_t \cdot I_{PS}(i,j,t) .
\end{eqnarray}
If the $PS$ exceeds a certain threshold $\xi_{PS}$ (e.g., 0.7), nodes $i$ and $j$ are deemed suitable for pairing and thus can be grouped together.
The FL server calculates $PS$ for every pair of nodes during the clustering period.
Once the suitability between all node pairs has been determined, the `\textit{Grouping Suitability Graph}' $G=(\tilde{N},E)$ can be formed where an edge set $E$ is obtained by adding an edge between nodes $i$ and $j$ if their $PS$ exceeds $\xi_{PS}$, while forcing them unconnected otherwise.

The above procedure is proposed as the Dynamic Clustering (DC) algorithm, whose pseudo-code is shown in Algorithm~\ref{alg:DC}. 

\begin{algorithm}[!t]
\caption{Dynamic Clustering (DC)}
\label{alg:DC}
\begin{algorithmic}[1]
\Statex \textbf{Input:} node set $\{ 1, \ldots, N \}$,
\Statex \quad \quad \quad $CS$ threshold $\xi_{CS}$, $PS$ threshold $\xi_{PS}$
\Statex \textbf{Output:} Node set $\tilde{N}$, Edge set $E$
\Statex /* At the \textit{clustering period} */
\For{\texttt{$i \in \{ 1, \ldots, N \}$}}
    \State $CS(i)$ $\gets$ \textit{Clustering Suitability} ($i$)
    \If{$CS(i) \geq \xi_{CS}$}
      \State node set $\tilde{N} \gets$ $i$
    \EndIf
\EndFor
\For{all pair of nodes $i, j \in \{ 1, \ldots, \tilde{N} \}$}
    \State $PS(i,j)$ $\gets$ \textit{Pairing Suitability ($i$, $j$)}
    \If{$PS(i,j) \geq \xi_{PS}$}
      \State edge set $E \gets$ edge between nodes $i$ and $j$
    \EndIf
\EndFor
\end{algorithmic}
\end{algorithm}

\subsection{Proposed Grouping Algorithm: Partial-Steady Grouping}

We can now apply the grouping technique to the obtained $G$.
In graph theory, two types of grouping techniques are most popular: \textit{clique partitioning} and \textit{graph coloring}.
Clique partitioning tries to divide a graph into separate and independent cliques \cite{Bhasker1991CMA}, while graph coloring aims to color the vertices of a graph such that no two adjacent vertices share the same color thereby forming distinct groups \cite{Lewis2015Springer}.
Then, it is well known that the partitioning of $G$ into cliques is equivalent to the coloring of its complement graph $G^{\prime}$ \cite{Bhasker1991CMA}. 
Since the density of $G^{\prime}$ is usually sparser than that of $G$ under the assumption of uniformly distributed nodes in the cluster,\footnote{$G$ assigns an edge to a pair of nodes if they are more than $d_{min}$ apart, whereas $G^{\prime}$ does so if they are closer than $d_{min}$. Since usually $d_{max} \gg d_{min}$, $G$ becomes much denser than $G^{\prime}$.} this work adopts the graph coloring method on $G^{\prime}$ in the sequel, to alleviate the computational complexity of grouping.

\subsubsection{Design considerations} 
We only consider `proper' coloring, i.e., coloring with no clashes (a clash refers to same-colored adjacent nodes \cite{Lewis2015Springer}).
If improper coloring is performed on $G^{\prime}$, nearby nodes could be grouped together resulting in potential inter-node data correlation.
We thus ruled out any known improper coloring techniques in our design.

Our main coloring objectives are: `even' grouping and minimizing the number of ungrouped nodes. 
Even grouping implies the group size (in terms of the number of nodes) should be as even as possible, to maintain a similar amount of per-group federated dataset leading to the consistent model performance across the groups.
This is to ensure little fluctuation in the global model evolution throughout successive rounds, regardless of the order of groups.
%
Next, it is also desirable to minimize the number of non-participating (i.e., ungrouped) nodes, not only to avoid free-riders in the FL training \cite{Lyu2020TPDS} but also to mitigate the training bias due to the excluded local data from the dropped devices \cite{Horvath2021Neurips}.
%
Therefore, the major challenge lies in jointly pursuing evenness and minimal ungrouped nodes with a right balance, within the scope of proper coloring.

Moreover, the reduction in the number of groups should also be considered because too many small-sized groups may incur a large latency in completing all the rounds.
%
In such a vein, it has been a main issue in graph coloring to search for the minimum number of colors $k$ to color a given graph, which is called the \textit{chromatic number}.
Since determining the chromatic number is NP-hard \cite{Garey1979Freeman}, numerous heuristics and metaheuristic approaches have been devised.
Among them, we adopt 
a popular metaheuristic method called 
\textit{PartialCol} \cite{Blochliger2008Computers} and 
revise it to best fit into the nature of our problem. 

Original PartialCol is 
designed to ascertain the chromatic number of a given graph,
by iteratively reducing $k$ from its initial value while maintaining the properness of coloring.
The following is an overview of the algorithm.
\begin{itemize}
\item At each iteration, the algorithm first constructs an initial solution $\{S_1, S_2,...S_k, U\}$ comprising $k$ color sets and one uncolored set $U$, via greedy coloring \cite{Lewis2015Springer} as follows.
    Let $P$ be a permutation of the nodes in $\tilde{N}$.
    For each node in $P$, the node is assigned to ${S_1}$ if the color set includes no adjacent nodes of the node. Otherwise, the node tries other color sets in the increasing order of the set index.
    If the node fails to find its color set, it is assigned to $U$. 
\item If the set $U$ is not an empty set, then the algorithm performs \textit{Tabu Search} (TS) on the initial solution.
    To empty $U$, an arbitrary node $i$ in $U$ is moved to a randomly selected set $S_j$.
    Then, the nodes in $S_j$ adjacent to $i$ are moved to $U$ and marked as `tabu', prohibiting their move back to $S_j$ for a pre-defined period.
\item TS continues to try the next move either until $|U|$ is successfully reduced to $0$ or until the maximum move count is reached.
    Such rules facilitate the exploration of unscouted solution spaces, increasing the chance of discovering an optimal solution that minimizes $|U|$.
\item If TS successfully reduces 
$|U|$ to $0$, the solution is saved and the algorithm 
restarts with $k$ decremented by one.
\end{itemize}

\subsubsection{Partial-Steady Grouping (PSG)}
\label{subsubsec:PSG}
Our PSG algorithm redefines the cost function of PartialCol (which was the number of ungrouped nodes $|U|$) into a joint cost as follows:
\begin{eqnarray}
\label{eq:alpha}
C = \alpha \cdot |U| + (1-\alpha) \cdot v , \quad 0 < \alpha < 1 ,
\end{eqnarray}
by combining $|U|$ with the variance $v$ of per-group sizes, where $v$ is a measure of evenness (smaller $v$ means better evenness) and $\alpha$ is a weighting factor between the two metrics.

Since PartialCol iteratively decreases $k$, the initial $k$ should be carefully chosen.
We propose to determine the initial $k$ by utilizing DSatur, a popular method for generating an initial solution in metaheuristics thanks to its ability to rapidly determine $k$ and build a proper solution \cite{Brelaz1979CACM}.
Although DSatur is known to produce $k$ close to the chromatic number, our algorithm can further decrease $k$ by allowing some nodes to be possibly uncolored for the sake of decreasing $k$, unlike DSatur that tries to find minimal $k$ with no dropout nodes.

Once $k$ is given, we need to construct an initial coloring solution, for which we propose a modified greedy algorithm called \textit{Equitable Largest-degree First} (ELF).
When coloring a node, \textit{Equitable} coloring \cite{Lewis2015Springer} tries to select the color set with the minimum number of nodes among possible proper sets, thus pursuing evenness.
In addition, \textit{Largest-degree First} coloring \cite{Gandhi2015ICCNC} determines the sequence of node coloring by prioritizing nodes according to their `adjacent node count' in the descending order.
By handling the node with the highest degree first, it intuitively leads to a smaller number of colors necessary to color a graph completely (or equivalently, it leads to a reduced number of uncolored nodes when the number of colors is limited).
Therefore, by combining the two, we can obtain a superior initialization for TS, so as to eventually derive a better final solution.

ELF determines its solution $s_{ELF} = \{S_1, S_2, \ldots, S_k, U\}$ similarly to the original PartialCol as follows.
First, $S_1, \ldots, S_k$ and $U$ start as empty sets.
Then, for given $G^{\prime}$, the set $P$ of the node coloring sequence is constructed by the \textit{Largest-degree First} rule.
Then, each node in $P$ is assigned to a color set according to the \textit{Equitable} coloring rule.
If no color set is available for the node, it is assigned to $U$.
After all nodes are assigned to $s_{ELF}$, the algorithm calculates the cost $C_{ELF}$ of the solution $s_{ELF}$ according to Eq.~\eqref{eq:alpha}.

Meanwhile, due to the updated cost function, the principle of TS should be modified such as: \textit{TS should continue its operation even if $|U|=0$ has been reached, as long as $C$ remains large}.
Specifically, the modified TS selects the color set  with the maximum number of nodes (denoted by $S_M$), and randomly move $r$ nodes in $S_M$ to $U$, where $r = |S_M| - |S_m|$ and $S_m$ is the color set with the minimum node count.
Then, thus-moved nodes are prohibited to move back to their original color set.
In the mean time, for the sake of evenness, the modified TS is designed to always reach the maximum iterations as it scouts for better solutions.
Other than the aforementioned changes, the modified TS is performed the same as the original TS, but now with three inputs: $s_{ELF}$, $G^{\prime}$, $\alpha$.
The modified TS finally yields its solution $s_{TS}$ as the one with the minimum $C$ throughout all the iterations.

Finally, we introduce `cos$t$-enhancing $r$atio' $tr$, to reduce the number of groups as much as possible.
Original PartialCol reiterates itself by decrementing $k$ by $1$ when reaching $|U|=0$, since it is no more possible to enhance the cost with given $k$.
In our case, however, $C=0$ is achieved extremely rarely due to the variance in the joint cost.
Hence, in contrast to the reiteration mechansim of original PartialCol, we propose the following procedure for reiteration.
According to the modified TS, if $C \leq C_p \times tr$ (where $C_p$ is the cost in the previous iteration), the algorithm decrements $k$ by 1 and reiterates the grouping mechanism.
That is, we treat `enhancing the previous cost $C_p$ by a certain fraction (i.e., $tr$)' as an indication for justifying the necessity of further decrementing $k$.
On the contrary, if $C > C_p \times tr$, PSG stops its operations and returns the previous solution (denoted by $s_p$) and its cost $C_p$.


The pseudo-code of PSG is provided in Algorithm~\ref{alg:PSG}.

\begin{algorithm}[!t]
\caption{Partial-Steady Grouping (PSG)}
\label{alg:PSG}
\begin{algorithmic}[1]
\Statex \textbf{Input:} complement graph \textit{$G'$}, $tr$, $\alpha$
\Statex \textbf{Output:} solution $k$, $s$, $C$
\Statex /* Right after running DC */
\State $k \gets$ \textit{DSatur} $(G')$
\State $\{s_{ELF}, C_{ELF}\}$ $\gets$ \textit{ELF greedy algorithm} ($G'$, $k$, $\alpha$)
\State $\{s_{TS}, C_{TS}\} \gets$ modified TS ($s_{ELF}$, $G^{\prime}$, $\alpha$)
\While{true}
  \State $k \gets k-1$
  \State $s_p \gets s_{TS}$ \Comment{\textit{store previous solution}}
  \State $C_p \gets C_{TS}$ \Comment{\textit{store previous cost}}
  \State $\{s_{ELF}, C_{ELF}\}$ $\gets$ \textit{ELF greedy algorithm} ($G'$, $k$, $\alpha$)
  \State $\{s_{TS}, C_{TS}\} \gets$ modified TS ($s_{ELF}$, $G^{\prime}$, $\alpha$)
  \If{$C_{TS} \leq C_p \times tr$}
   \State continue \Comment{\textit{try to reduce $k$ further}}
  \Else \Comment{\textit{no cost improvement}}
   \State $k \gets k+1$ \qquad $s \gets s_p$ \qquad $C \gets C_p$ 
   \State \Return \Comment{\textit{return the previous results}}
  \EndIf
\EndWhile
\end{algorithmic}
\end{algorithm}

\subsection{Overall Procedure of the Proposed Mechanism}

Our proposed FL mechanism is summarized as follows.
The FL server executes DC in Algorithm~\ref{alg:DC} based on the location history of $N$ nodes for the last $T$ time samples, 
obtaining the Grouping Suitability Graph $G$.
Subsequently, the server applies PSG in Algorithm~\ref{alg:PSG} to $G^{\prime}$, and 
$k$ independent groups are formed upon completion of PSG, each of which will participate in FL during its own corresponding round.

In practice, the FL server should be able to complete the DC and PSG algorithms within the clustering period.
Since the majority of the execution time is dedicated to the modified TS, we propose a stopping condition to terminate the execution of the modified TS, to reduce its runtime.
We refer to our proposed stopping method as \textit{REvES} (Runtime Enhancement via Early Stopping), which works as follows.
In the modified TS, REvES observes the minimum and maximum $C$ within the moving window of size $ws$ (iterations), where the window shifts by one at each iteration.
If both of the minimum and maximum $C$ remains constant for a consecutive period of $p$ iterations, the modified TS terminates itself and returns the best solution with the smallest $C$.
Note that the effect of REvES will be further analyzed in Section~\ref{sec:eval}. 

\section{Performance Evaluation}
\label{sec:eval}
This section evaluates our proposed algorithms via simulations across various scenarios.
Then, our grouping algorithm's performance is compared with three benchmark coloring algorithms, followed by an analysis on the effect of REvES.

\subsection{Scenario Setup}

Table~\ref{tab:scenariospec} shows the three scenarios considered, densely-, moderately-, and sparsely-populated areas, with device population density (in $/m^2$) of $10^{-1}, 10^{-2}, 10^{-3}$, respectively.\footnote{We referred to the human population density in \cite{Deville2014PNAC, Petrasova2019ISPRS}, assuming that each person possesses an IoT device like mobile handset.}
The dense case's device density is set based on the data in a public square in a city \cite{Petrasova2019ISPRS}, the sparse case's density is obtained from the data taken in a sub-urban area \cite{Deville2014PNAC}, and the moderate case's density is set as a mid-point between the two aforementioned cases.
Then, we also consider `FL-participating device density', which is defined as the device population density multiplied by the mobile application popularity participating in the FL process (e.g., Instagram's app popularity: 40\% \cite{Brooke2021website}).
In addition, we assumed that each scenario's area is a square with $100$, $200$, and $1000$ meters per side, respectively.

\begin{table}[!t]
\resizebox{\columnwidth}{!}{%
\begin{tabular}{|c|c|c|c|}
\hline
 & densely- & moderately- & sparsely- \\
 & populated & populated & populated \\
\hline
device population density              & $10^{-1} /m^2$         & $10^{-2} /m^2$         & $10^{-3} /m^2$         \\ \hline
FL-participating device & $4\times 10^{-2} /m^2$ & $4\times 10^{-3} /m^2$ & $4\times 10^{-4} /m^2$ \\
density (40\% popularity) & & & \\ \hline
area size                      & $100 \times 100$ m$^2$     & $200 \times 200$ m$^2$     & $1 \times 1$ km$^2$       \\ \hline
$d_{min}$, $d_{max}$            & 10m, 100m              & 32m, 200m              & 100m, 1000m            \\ \hline
\end{tabular}
}
\caption{Simulation parameters for the three tested scenarios}
\label{tab:scenariospec}
\end{table}

Mobile devices are assumed uniformly distributed at random locations following the Poisson point process, each moving to a random direction at a random speed comparable to the human walking speed.
We also assume that an FL server's cluster is circular with the diameter of $d_{max}$ which is inscribed in the square-shaped area of each scenario.\footnote{This implies we have assumed that a less-populated scenario has a larger area and bigger $d_{max}$. For example, an urban riverside park may occupy a quite large area with relatively consistent device density and environment.}
%
Furthermore, we set $d_{min} = 10$m for the densely-populated scenario, while $d_{min}$ in the other two scenarios is set as
$10$m $\times \sqrt{ {\scriptstyle {10^{-1}}/{10^{-2}}} } \simeq 32$m for the moderately-populated scenario and
$10$m $\times \sqrt{ {\scriptstyle {10^{-1}}/{10^{-3}}} } = 100$m for the sparsely-populated scenario.
%
Such density-driven setting of $d_{min}$ has been motivated by the situations where the inter-device dependency in their IoT data is affected by the device density, such as object detection, speech recognition, etc.
Note that the square root has been applied since $d_{min}$ is one-dimensional whereas the density is two-dimensional.


\subsection{Ablation Study on PSG Algorithm}

We first conducted an ablation study to reveal the individual impact of each design component in the PSG algorithm.
The study compares original PartialCol with one of its evolved versions, where the latter is obtained by selectively applying ELF or modified TS (with two $tr$ values) to the original PartialCol.
As performance metrics for comparison, we measured `the number of groups' and `joint cost', for the three test scenarios. 
For each scenario, 20 realizations of node deployment have been generated for the DC algorithm, and for each realization each version of grouping algorithm (i.e., PartialCol, PartialCol with ELF, PartialCol with modified TS) has been executed 20 times.
The average of thus-obtained 400 results is recorded in Table~\ref{tab:AblationStudy}.

\begin{table}[!t]
\centering
\resizebox{\columnwidth}{!}{%
\Large{
\begin{tabular}{@{}c|cccccc@{}}
\toprule[1.5pt]
 & \multicolumn{2}{c}{\textbf{densely-populated}} & \multicolumn{2}{c}{\textbf{moderately-populated}} & \multicolumn{2}{c}{\textbf{sparsely-populated}} \\
 & groups & cost($\alpha=0.5$) & groups & cost($\alpha=0.5$) & groups & cost($\alpha=0.5$) \\ \midrule
PartialCol              & 9            & 292.29      & 9.55       & 26.87       & 11.5          & 141.02 \\ \midrule
\multirow{2}{*}{+ ELF}  & 9       & 62.42 & 9.55      & 8.80   & 11.50 & 32.08 \\
                        & (0\%)       & (-78.6\%) & (0\%)      & (-67.3\%)   & (+0.02\%) & (-77.3\%) \\ \midrule
+ mod. TS & 9.5 & 29.03 & 10.38 & 0.1131 & 12.21 & 21.10 \\
($tr=0.7$) & (+5.56\%) & (-90.1\%) & (+8.69\%) & (-95.9\%) & (+6.20\%) & (-85\%) \\ \midrule
+ mod. TS & 9.6 & 37.72 & 10.49 & 0.2057 & 12.30 & 22.37 \\
($tr=0.1$) & (+6.67\%) & (-87.1\%) & (+9.87\%) & (-94.8\%) & (+6.89\%) & (-84.1\%) \\ \bottomrule[1.5pt]
\end{tabular}%
}}
\caption{Ablation study results}
\label{tab:AblationStudy}
\end{table}

\subsubsection{Impact of ELF}
As shown in Table~\ref{tab:AblationStudy}, applying ELF to PartialCol's initialization decreases the cost by at least 67.3\% and up to 78.6\%, without sacrificing the number of groups at all.
Since both PartialCol and `PartialCol + ELF' have no ungrouped nodes in their solutions, we set their cost as a half of the evenness (i.e., $\alpha=0.5$).
As intended, enhancing the initial solution of PartialCol by adopting ELF produces a superior solution indeed.

\subsubsection{Impact of $C$, modified TS, and $tr$}
In case the original PartialCol's cost is revised to the joint cost $C$, both $TS$ and the strategy of reducing the number of groups should be adjusted as well (as discussed in Section \ref{subsubsec:PSG}).
Accordingly, we analyzed the effect of revising the cost function and TS, and that of introducing $tr$ (with two different values, $0.7$ and $0.1$), while setting $\alpha=0.5$ due to the absence of ungrouped nodes in the original PartialCol.
Table~\ref{tab:AblationStudy} presents that `PartialCol + modified TS' with $tr=0.7$ can mitigate $C$ significantly by 90.3\% on average (averaged over the three scenarios) while mildly increasing the number of groups only by 6.81\% on average, compared to the original PartialCol.
With smaller $tr$ (i.e., $tr=0.1$), the performance of `PartialCol + modified TS' still outperforms the original PartialCol, with a slight degradation from the case of $tr=0.7$.

\subsection{Performance of the Proposed Mechanism}

We set our algorithm parameters as follows.
For the clustering algorithm, the thresholds $\xi_{CS}$ and $\xi_{PS}$ are both set to $0.7$, and the discount factor in $CS$ and $PS$ follows a linear monotonic increasing function such as $\omega_t = {t}/{\sum_{t = 1}^\textit{T} t}$.
For each scenario, we have generated 20 realizations of node deployment for the DC algorithm, and for each realization we have run the PSG algorithm 20 times.
Then, we took the average of thus-obtained 400 results, for each of the following four metrics we measured: the number of groups, the joint cost, the number of ungrouped nodes, and evenness.
In addition, for PSG, $tr$ varied as $0.7$, $0.4$, $0.1$.

\begin{figure}[!t]
    \centering
    \begin{subfigure}{\columnwidth}
        \centering
        \includegraphics[width=\columnwidth]{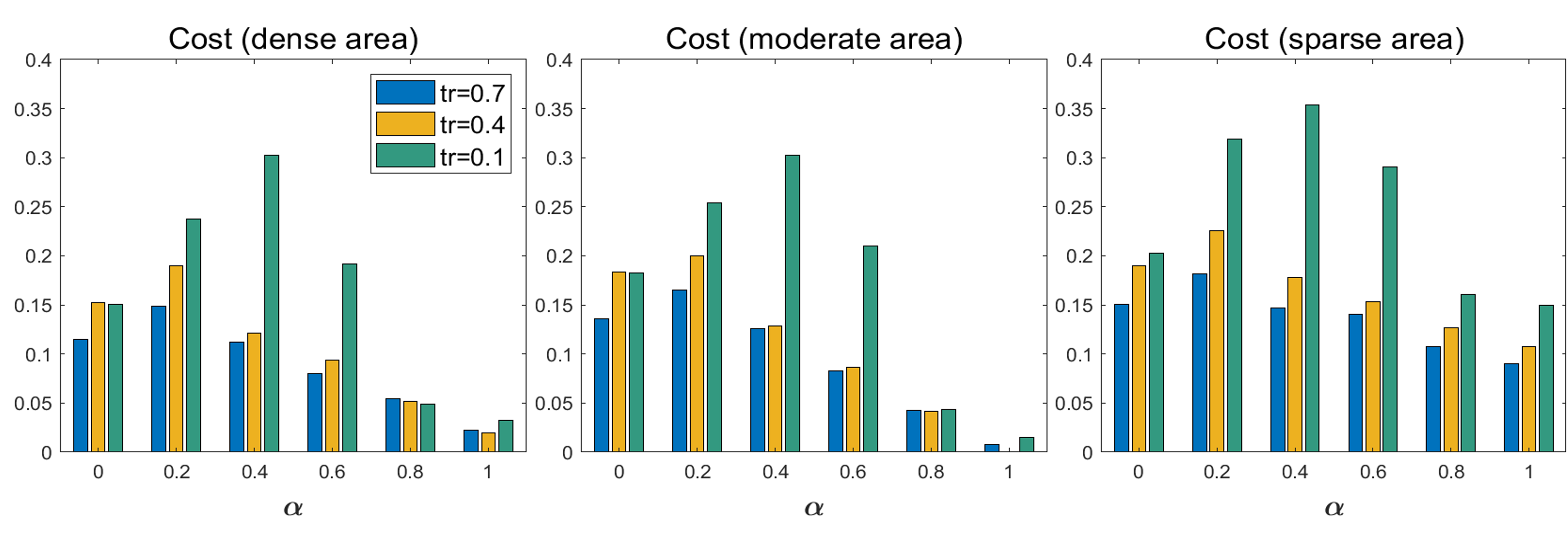}
        \vspace{-0.25in}
        \caption{joint cost (densely- vs. moderately- vs. sparsely-populated)}
	    \label{fig:simulresult_a}
	    \vspace{0.1in}
    \end{subfigure}
    \begin{subfigure}{\columnwidth}
        \centering
        \includegraphics[width=0.8\columnwidth]{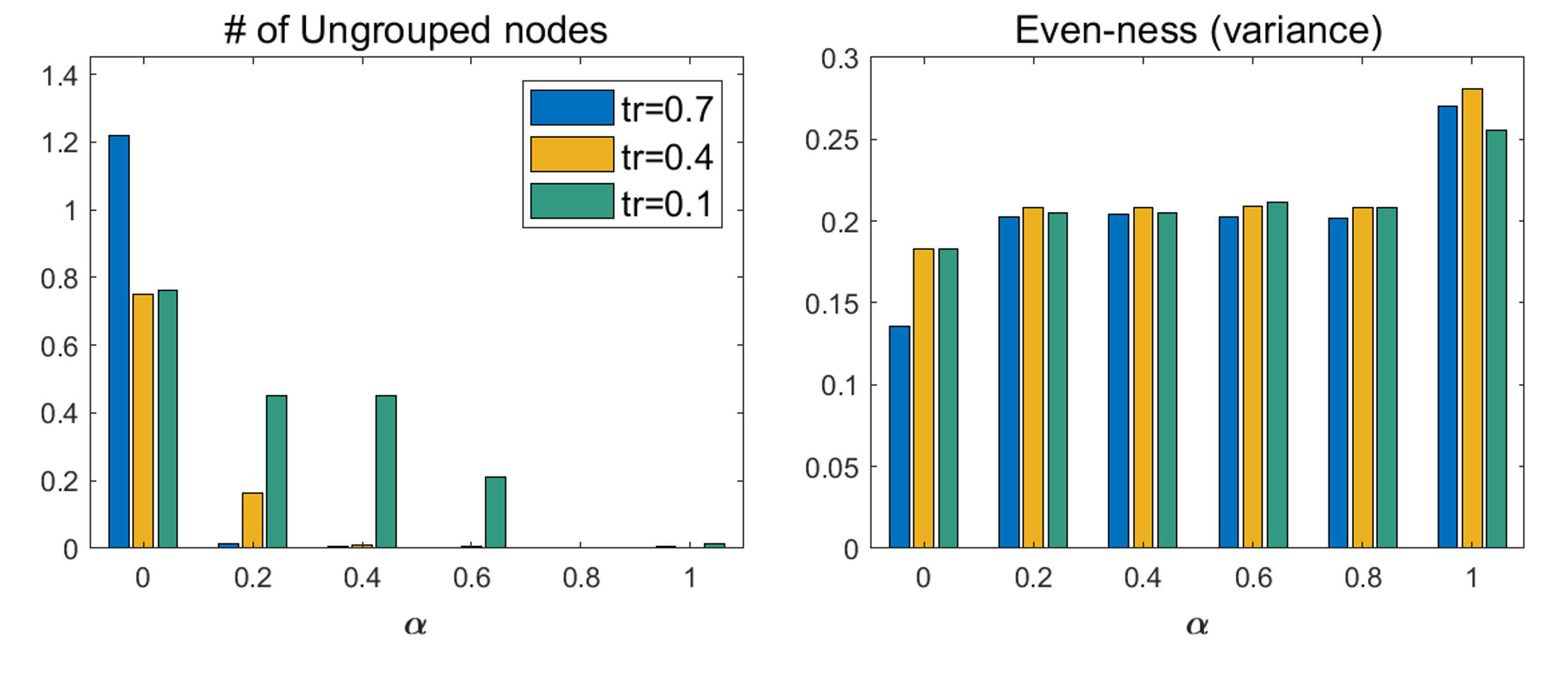}
        	\vspace{-0.1in}
        \caption{`\# of ungrouped nodes' and evenness (for moderately-populated)}
	    \label{fig:simulresult_b}
    \end{subfigure}
    %
    %
    %
    \caption{The simulation results of the PSG algorithm for three tested environments ($\alpha$ and $tr$: defined in Section~\ref{subsubsec:PSG})} 
    \label{fig:simulresult}
\end{figure}
%

\begin{table}[!t]
\centering
\resizebox{\columnwidth}{!}{%
\large{
\begin{tabular}{@{}c|cccccc@{}}
\toprule[1.5pt]
 & \multicolumn{2}{c}{\textbf{densely-populated}} & \multicolumn{2}{c}{\textbf{moderately-populated}} & \multicolumn{2}{c}{\textbf{sparsely-populated}} \\
 & groups & cost($\alpha=0.5$) & groups & cost($\alpha=0.5$) & groups & cost($\alpha=0.5$) \\ \midrule
PSG($tr=0.7$) & 9.575 & 0.0953 & 10.44 & 0.1029 & 12.283 & 0.1475 \\
PSG($tr=0.4$) & 9.6 & 0.1013 & 10.48 & 0.1131 & 12.30 & 0.1543 \\
PSG($tr=0.1$) & 9.6 & 0.2251 & 10.47 & 0.2057 & 12.3 & 0.2832 \\ \midrule
DSatur & 9.3 & 287.83 & 9.95 & 25.47 & 11.9 & 148.14 \\
PartialCol & 9 & 292.29 & 9.55 & 26.87 & 11.5 & 141.02 \\
TabuCol & 9 & 264.95 & 9.56 & 22.74 & 11.51 & 129.83 \\ \bottomrule[1.5pt]
\end{tabular}%
}}
\caption{Performance comparison in node grouping}
\label{tab:AlgComparison}
\end{table}
%

Fig.~\ref{fig:simulresult} and the upper three rows of Table~\ref{tab:AlgComparison} present the results of the PSG algorithm across the three scenarios, where Fig.~\ref{fig:simulresult} varies $\alpha$ from $0$ to $1$ at intervals of $0.2$ while Table~\ref{tab:AlgComparison} sets $\alpha=0.5$.
Table~\ref{tab:AlgComparison} shows that `the number of groups' tends to gradually increase as the environment becomes sparser, whereas it is insensitive to $tr$.
The `joint cost' achieved by PSG, however, is affected by both device density and $tr$ as shown in Table~\ref{tab:AlgComparison} and Fig.~\ref{fig:simulresult_a}, where the cost tends to deteriorate as device density or $tr$ decreases.
%
Regarding `the number of ungrouped nodes' and `evenness', however, we found that the three scenarios share similar tendency, and hence we focused on the moderately-populated case in Fig.~\ref{fig:simulresult_b} to show the tendency of the two metrics with varying $tr$ and $\alpha$.
As shown, with the increase of $\alpha$, the number of ungrouped nodes decreases whereas the evenness deteriorates, since the joint cost $C$ puts more weight on $|U|$ as $\alpha$ approaches 1.
%

\subsection{Performance Comparison with Baseline Algorithms}

Our PSG algorithm has been compared with three state-of-the-art algorithms, PartialCol \cite{Blochliger2008Computers}, TabuCol \cite{Hertz1987Computing}, and DSatur \cite{Brelaz1979CACM}.
Similar to PartialCol, TabuCol is a metaheuristic utilizing Tabu Search to minimize the initial solution's cost with the goal of finding the minimum number of groups.
In TabuCol, however, the cost is defined as the count of clashes within the current solution, considering TabuCol's strategy of assigning a node to a random color set when no appropriate color set is available for proper coloring.
DSatur is a polynomial-time algorithm in metaheuristic coloring, which is popularly used thanks to its quick generation of feasible initial solutions.
The algorithm selects the node with the largest \textit{saturation degree} (i.e., the number of different colors used by its neighbors) to color with the minimal-indexed color.
In doing so, if a node cannot be colored properly, $k$ is increased by $1$.

Since the aforementioned algorithms always yield proper solutions (i.e., no ungrouped nodes in the solution), we measured the evenness of their solutions (by following our definition of evenness) and took its half, to compare with our algorithm's cost with $\alpha=0.5$.
As shown in Table~\ref{tab:AlgComparison}, our cost is significantly smaller than the benchmark algorithms in every scenario, with a slight sacrifice in the number of groups.
Specifically, the greatest disparity in the number of groups between the benchmarks and ours is only 0.93 (in the moderately-populated scenario), while the achieved cost is at least 110 times smaller than that of the counterparts.
That is, our proposed algorithm shows significant enhancement in evenness and the number of ungrouped nodes, with minimal compromise in the group count.

\subsection{Impact of REvES in Grouping}

Recall that REvES aims to minimize the number of iterations run by our modified TS.
We examined whether PSG with REvES can notably reduce the number of iterations without severely increasing the cost and the number of groups.
With $tr=0.7$, PSG with REvES has been simulated for the three population scenarios while varying $\alpha$ from 0.1 to 0.9, 
and the nine results obtained by varying $\alpha$ are averaged.
Then, we utilized thus-derived values (in terms of
the number of groups, cost, and the total number of iterations) to calculate how much change `PSG with REvES' incurs compared to `PSG without REvES'.
For REvES, we set $ws=150$ and $p=70$.

\begin{table}[!t]
\centering
\resizebox{\columnwidth}{!}{%
\begin{tabular}{@{}cccc@{}}
\toprule
Comparison categories & densely-populated & moderately-populated & sparsely-populated \\ \midrule
\multicolumn{1}{c|}{change in \# of groups} & 0.084\% & 0.157\% & 0.020\% \\
\multicolumn{1}{c|}{change in cost} & 0.323\% & 14.08\% & 8.387\% \\
\multicolumn{1}{c|}{change in total iterations} & -73.67\% & -70.87\% & -73.51\% \\ \bottomrule
\end{tabular}%
}
\caption{Performance of `PSG with REvES' against `PSG without REvES' (when $tr=0.7$)}
\label{tab:REvES_result}
\end{table}

As indicated in Table~\ref{tab:REvES_result}, REvES reduces the total number of iterations in PSG by 72.85\% on average.
Concurrently, the number of groups remains nearly unchanged, and the average cost mildly increases (by 0.32\% $\sim$ 14.08\%). 
Due to its ability to substantially decrease the iteration count without causing a detrimental cost escalation, REvES could find its utility in time-intensive scenarios, e.g., with many nodes involved.

\section{Conclusion}
\label{sec:conclusion}
This paper proposed a geographical approach to alleviating the non-IID data problem in FL, based on our intuition on the inter-device distance driven IoT data independence and identicalness.
Facilitated by graph coloring methods, we have developed node clustering and grouping algorithms to achieve an almost-IID dataset per FL group.
Through extensive simulations, 
we demonstrated that our mechanism much outperforms the existing alternatives in varying IoT environments.

In the future, we will investigate the general impact of the proposed methods by considering more diversified IoT data types (e.g., vision, audio) and environments (e.g., outdoor). In addition, we will try to deploy our proposed mechanism with existing FL algorithms to show its efficacy in real applications.

\bibliographystyle{IEEEtran}
\bibliography{references}

\end{document}